\documentclass[conference]{ieeeconf}
\pdfoutput=1
\IEEEoverridecommandlockouts

\usepackage[backend=biber,
            hyperref=true,
            url=false,
            isbn=false,
            doi=false,
            backref=false,
            style=ieee,
            citestyle=numeric-comp,
            sorting=nyt,
            block=none]{biblatex}
        
\renewcommand{\bibfont}{\small}
\usepackage{flushend}
\usepackage{balance}
\usepackage{marginnote}
\usepackage{graphics}
\usepackage[pdftex]{graphicx}
\usepackage{wrapfig}
\DeclareGraphicsExtensions{.pdf,.png,.jpg}

\usepackage[font={small}]{caption}
\usepackage{subcaption}
\usepackage[rightcaption]{sidecap}
\usepackage{pbox}

\usepackage{mathtools}
\usepackage{amsmath, amssymb, amscd}
\usepackage{ wasysym } 
\usepackage{amsfonts}
\usepackage{mathptmx} 
\usepackage{gensymb} 
\usepackage{nicefrac}       

\DeclareMathAlphabet{\mathcal}{OMS}{lmsy}{m}{n}
\DeclareSymbolFont{largesymbols}{OMX}{cmex}{m}{n}
\usepackage{textcomp} 
\usepackage{dsfont}

\usepackage{array} 
\usepackage{tabularx}
\usepackage{multirow}
\usepackage{multicol}
\usepackage{booktabs}

\usepackage[
  separate-uncertainty = true,
  multi-part-units = repeat
]{siunitx}
\usepackage[T1]{fontenc} 
\usepackage[utf8]{inputenc}
\usepackage[english]{babel} 
\usepackage{units}
\usepackage{bm}
\usepackage{times} 
\usepackage{xspace}
\usepackage{flushend}
\usepackage{csquotes}
\usepackage{makeidx}

\let\labelindent\relax
\usepackage{enumitem}

\usepackage{soul} 
\usepackage{subfiles} 

\usepackage[protrusion=true,expansion=true]{microtype}
\usepackage[yyyymmdd]{datetime}

\date{\protect\formatdate{1}{1}{2001}}

\usepackage{url}
\makeatletter
\g@addto@macro{\UrlBreaks}{\UrlOrds}
\makeatother
\usepackage{color}
\usepackage[usenames,dvipsnames,table]{xcolor}
\usepackage[pdfborder={0 0 0.5}]{hyperref}
\hypersetup{
    colorlinks=true,
    linkcolor=black,
    citecolor=black,
    filecolor=cyan,
    urlcolor=black
}

\usepackage{soul} 

\newcommand{\tocite}[1]{%
\textcolor{red}{[cite:\ifthenelse{\equal{#1}{}}{}{#1}?]}
}

\newcommand{\ignore}[1]{}

\usepackage{color,soul}

\usepackage{algorithm}
\usepackage{algpseudocode}

\newcommand{\algoName}{IWR\xspace}
\newcommand{\algoFull}{Intervention Weighted Regression\xspace}

\begin{document}

\title{\LARGE \bf Human-in-the-Loop Imitation Learning using Remote Teleoperation}

\author{%
Ajay Mandlekar$^{1}$,
Danfei Xu$^{*1}$,
Roberto Mart\'in-Mart\'in$^{*1}$, 
Yuke Zhu$^{2}$,
Li Fei-Fei$^{1}$,
Silvio Savarese$^{1}$

\thanks{$^{*}\,$These authors contributed equally.$^{1}\,$Stanford Vision \& Learning Lab, $^{2}\,$The University of Texas at Austin.
}%
}

\maketitle

\begin{abstract}
Imitation Learning is a promising paradigm for learning complex robot manipulation skills by reproducing behavior from human demonstrations. 
However, manipulation tasks often contain \emph{bottleneck} regions that require a sequence of precise actions to make meaningful progress, such as a robot inserting a pod into a coffee machine to make coffee.
Trained policies can fail in these regions because small deviations in actions can lead the policy into states not covered by the demonstrations. 
Intervention-based policy learning is an alternative that can address this issue -- it allows human operators to monitor trained policies and take over control when they encounter failures.
In this paper, we build a data collection system tailored to 6-DoF manipulation settings, that enables remote human operators to monitor and intervene on trained policies.
We develop a simple and effective algorithm to train the policy iteratively on new data collected by the system that encourages the policy to learn how to traverse bottlenecks through the interventions. 
We demonstrate that agents trained on data collected by our intervention-based system and algorithm outperform agents trained on an equivalent number of samples collected by non-interventional demonstrators, and further show that our method outperforms multiple state-of-the-art baselines for learning from the human interventions on a challenging robot threading task and a coffee making task. 
Additional results and videos at \url{https://sites.google.com/stanford.edu/iwr}
\end{abstract}

\IEEEpeerreviewmaketitle
\section{Introduction}
\label{sec:intro}

Imitation Learning (IL) is a promising paradigm for learning complex manipulation skills by reproducing behaviors from human demonstrations~\cite{pomerleau1989alvinn, zhang2017deep, mandlekar2020learning}. Unlike interactive learning techniques, such as reinforcement learning, which generate large amounts of training data via autonomous exploration, the efficacy of IL is bounded by the cost of human demonstrations. This cost limits the amount of data available to train IL models.
Consequently, models trained by IL can suffer from covariate shift: small errors in actions can bring the learner to unseen states that the learner has not been trained for. To address this covariate shift problem, \textsc{DAgger}-style methods~\cite{ross2011reduction,laskey2016shiv,kelly2019hg} have an expert relabel dataset samples collected by the trained agent with actions that the expert would have taken. This allows training data to include samples that a trained agent is likely to encounter.


For real-world robotic tasks, however, \textsc{DAgger}-style data relabeling is often infeasible. For example, a 30-second manipulation task with 20 hz robot control would require a human to relabel 600 state samples for every trajectory collected by the robot. Moreover, the human needs to estimate the correct robot action that should have been taken in each state.
This kind of offline relabeling requires significant human effort and is prone to incorrect action labels~\cite{laskey2017comparing}.
Instead, it is more natural for humans to annotate actions \emph{in the loop}, i.e. monitor policy execution and take over control when intervention is needed~\cite{kelly2019hg, spencerlearning, ablett2020fighting}.

\begin{figure}[!t]
    \centering
    \includegraphics[width=\linewidth]{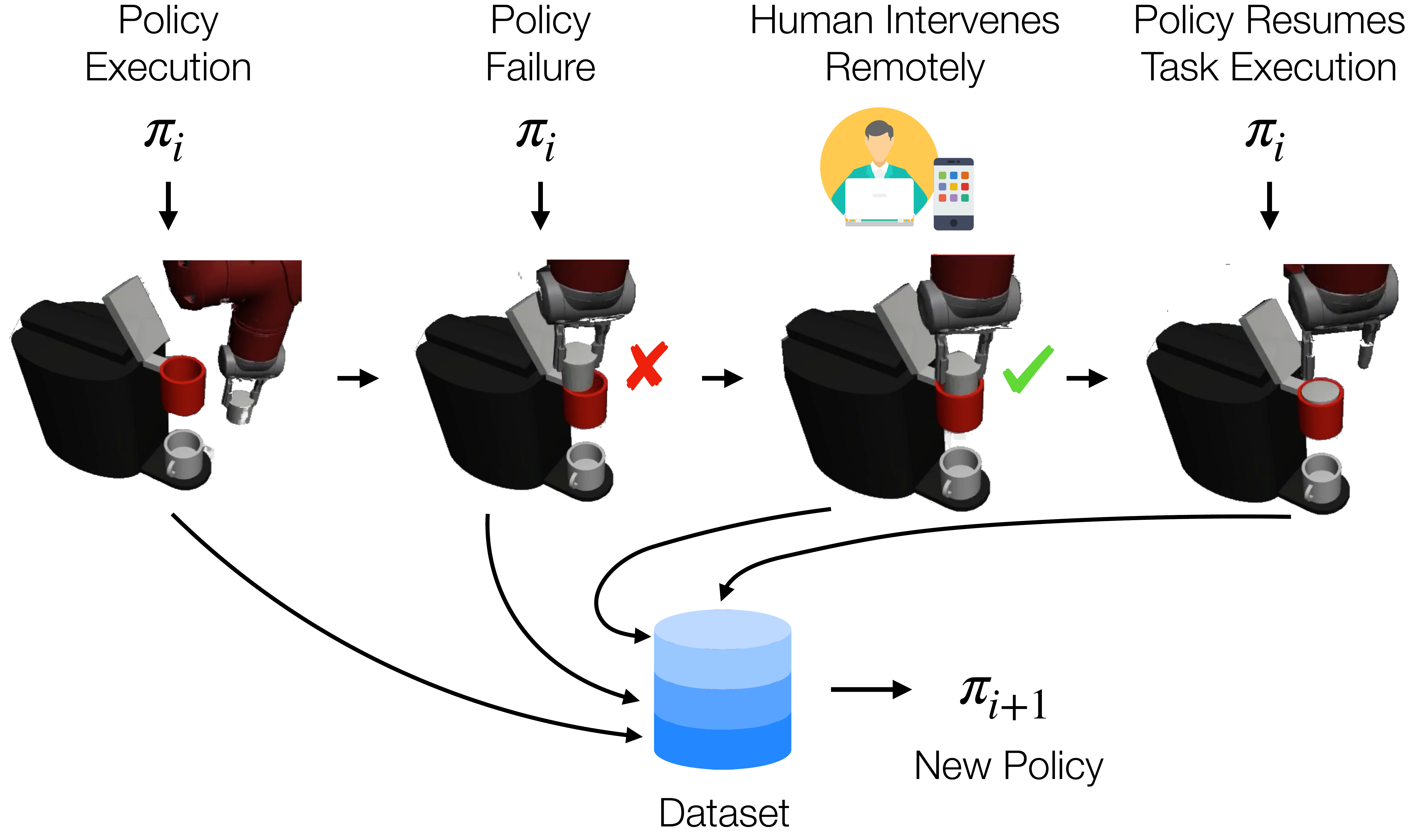}
    \caption{\textbf{Human-in-the-Loop Policy Learning with Human Interventions.} Manipulation tasks often contain bottleneck regions that require a series of precise actions to traverse successfully. Models trained on an offline set of human demonstrations easily fail in these regions due to action errors that compound. To address this issue, we built a system where a human operator observes a policy attempting to solve a manipulation task and intervenes when necessary to help solve the task. During an intervention, the human operator takes control of the arm from the policy, moves the robot arm into a state where the policy is likely to succeed, and then returns control to the policy. All data is aggregated into a dataset and the policy is re-trained on the new collected data. This process repeats. 
    }
    \label{fig:pull}
    \vspace{-5pt}
\end{figure}

However, intervention-based learning has mostly been limited to 2D driving domains~\cite{kelly2019hg, spencerlearning} where an agent must learn a policy to stay on the road. Both the data collection and the policy learning are straightforward in this domain -- humans can easily provide intervention actions in 2D, 
and the domain is tolerant to action error, since there is a large set of actions that keep the agent on the road. By contrast, in 6-DoF robot manipulation settings, certain regions of the state space can require precise sequences of actions to make meaningful task progress. 
These regions are much less tolerant to error, and a small deviation means the difference between success and failure. We call such regions \textit{bottlenecks}.

Consider the coffee making task shown in Fig.~\ref{fig:pull}, where the robot must carefully insert the pod into the machine slot. States where the pod is close to the container form a bottleneck, since only a particular sequence of actions lead to successful insertion and any deviation will cause the pod to collide with the rim. 
Tasks with such bottlenecks are ideal testbeds for intervention-based learning, because small inaccuracies in the output actions can make IL agents susceptible to making mistakes in these regions. 

Making human interventions feasible for robot manipulation raises technical challenges on the system side as well as on the algorithmic side. First, we need a robust system that enables human demonstrators to monitor the robot behaviors and gain immediate full control of the robot when observing imminent risks. 
Second, we need an effective algorithm to learn from human intervention data.
To this end, we develop a system suitable for collecting intervention data from remote users with 6-DoF control of robot end-effectors and a simple yet effective method to leverage the human interventions.

The key intuition behind our algorithm is that, in manipulation tasks, humans tend to intervene when the robot has difficulty ``entering'' a bottleneck and return control to the robot after traversing the bottleneck. Therefore, the human interventions are informative about both \textit{where} task bottlenecks occur and \textit{how} to traverse them. 
The algorithm we propose leverages these two signals for policy learning.
Specifically, we find that treating the intervention signal as an implicit reward function and performing weighted regression is highly effective to correctly leverage the new data.

\noindent \textbf{Summary of Contributions:}
\begin{enumerate}[
    topsep=0pt,
    noitemsep,
    leftmargin=*,
    itemindent=12pt]
\item We develop a system that enables remote teleoperation for 6-DoF robot control and a natural human intervention mechanism well suited to robot manipulation.
\item We introduce \algoFull (\algoName), a simple yet effective method to learn from human interventions that encourages the policy to learn how to traverse bottlenecks through the interventions. 
\item We evaluate our system and method on two challenging contact-rich manipulation tasks: a threading task and coffee machine task. We demonstrate that policies trained on data collected by our system outperform policies trained on an equivalent amount of full human demonstration trajectories, that \algoName outperforms alternatives for learning from the intervention data, and that our results hold across data collected from multiple human operators.
\end{enumerate}

\section{Related Work}
\label{sec:related}

\begin{figure}[!t]
    \centering
    \includegraphics[width=\linewidth]{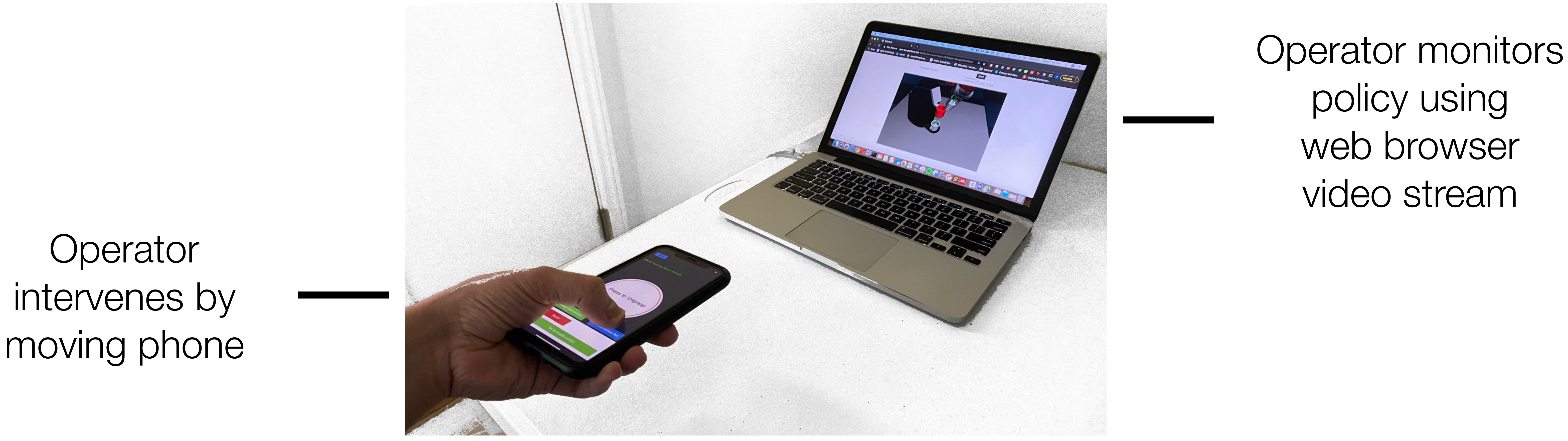}
    \caption{\textbf{System Overview.} We extend the RoboTurk system~\cite{mandlekar2018roboturk, mandlekar2019scaling} to enable remote users to monitor policy execution and intervene when they would like to take over control using a web browser and a smartphone. The user monitors policy execution by watching a video stream of the robot in their web browser. They can hold a button on their smartphone to intervene and control the robot by moving their smartphone.}
    \label{fig:system}
    \vspace{-10pt}
\end{figure}

\textbf{Imitation Learning from Offline Demonstrations:} Imitation learning can be used to train a policy from a set of offline expert demonstrations. A policy can be learned without any interaction, as in Behavioral Cloning (BC)~\cite{pomerleau1989alvinn, zhang2017deep, mandlekar2020learning} or using additional interaction with the environment, as in Inverse Reinforcement Learning (IRL)~\cite{abbeel2004apprenticeship, abbeel2011inverse}. However, policies trained with BC can suffer from \textit{covariate shift} because they are trained completely offline~\cite{ross2011reduction} -- it can be easy for the policy to encounter unseen states during evaluation. Because of this problem, BC often takes a prohibitive number of expert samples to work well. By contrast, IRL generally requires fewer expert samples, but it relies on a prohibitive amount of agent interaction with the environment~\cite{ho2016generative, zhu2018reinforcement, xu2019positive} due to the need to do reinforcement learning with a learned reward function. We instead focus on human-in-the-loop policy learning, which can strike a better balance between the number of human samples and agent samples required for learning.

\textbf{Human-in-the-Loop Policy Learning:} Human-in-the-Loop Policy Learning allows a human to provide additional supervision during the policy learning process. One paradigm is Reinforcement Learning (RL) with human feedback~\cite{zhang2019leveraging}, where a human provides rewards during agent training~\cite{loftin2016learning, macglashan2017interactive, cabi2019scaling, christiano2017deep, singh2019end, reddy2019learning}, but this suffers from the same limitations as IRL due to the need for extensive agent interaction.

\textsc{DAgger}~\cite{ross2011reduction} introduced a useful paradigm that can require less agent samples than IRL and less human samples than BC by asking an expert to relabel data collected by a trained agent with actions that should have been taken. 
However, \textsc{DAgger} is not feasible for humans in practical scenarios (e.g. continuous 6D control) due to the relabeling procedure, which can be burdensome and prone to human error, especially in manipulation settings~\cite{laskey2017comparing}. Prior work has attempted to reduce the number of human annotations needed~\cite{chernova2009interactive, laskey2016shiv, packard2017policies} but relabeling is still required. Noise injection during expert demonstrations has also been proposed in order to correct for covariate shift~\cite{laskey2017dart}. 
Other paradigms for human-in-the-loop policy learning include collaboration~\cite{hadfield2016cooperative, fisac2020pragmatic}, teaching the robot through informative sample selection~\cite{cakmak2012algorithmic, ho2016showing, brown2019machine}, and leveraging physical kinesthetic corrections~\cite{bajcsy2017learning, bajcsy2018learning}.

\textbf{Learning from Interventions:} In this work, we build a system that allows remote users to monitor policy execution and provide interventions, and a method that learns from the interventions intelligently. While similar systems have been built recently~\cite{ablett2020fighting, delpreto2020helping}, our system is the only one allowing for remote web-based operation and the only one demonstrated on contact-rich manipulation tasks.

Prior intervention-based approaches~\cite{kelly2019hg, ablett2020fighting} only leverage the human intervention samples for training the policy, and discard the agent samples that lead to those interventions. However, training without these on-policy agent samples can cause the behavior of the agent to change significantly after training on the new dataset, inducing a new distribution of mistakes due to covariate shift. By contrast, including the on-policy agent samples during training can help alleviate this issue by encouraging the agent to visit states where human interventions are available, increasing the likelihood of successful bottleneck traversal. 
Spencer et al.~\cite{spencerlearning} also found it useful to leverage all samples for learning, but their method only applies to discrete action settings and was only demonstrated in driving domains. 
Our method encourages successful bottleneck traversal in continuous control settings by re-weighting the dataset distribution to prioritize intervention samples over the on-policy samples, which are included for regularization.


\begin{figure}[!t]
    \centering
    \includegraphics[width=\linewidth]{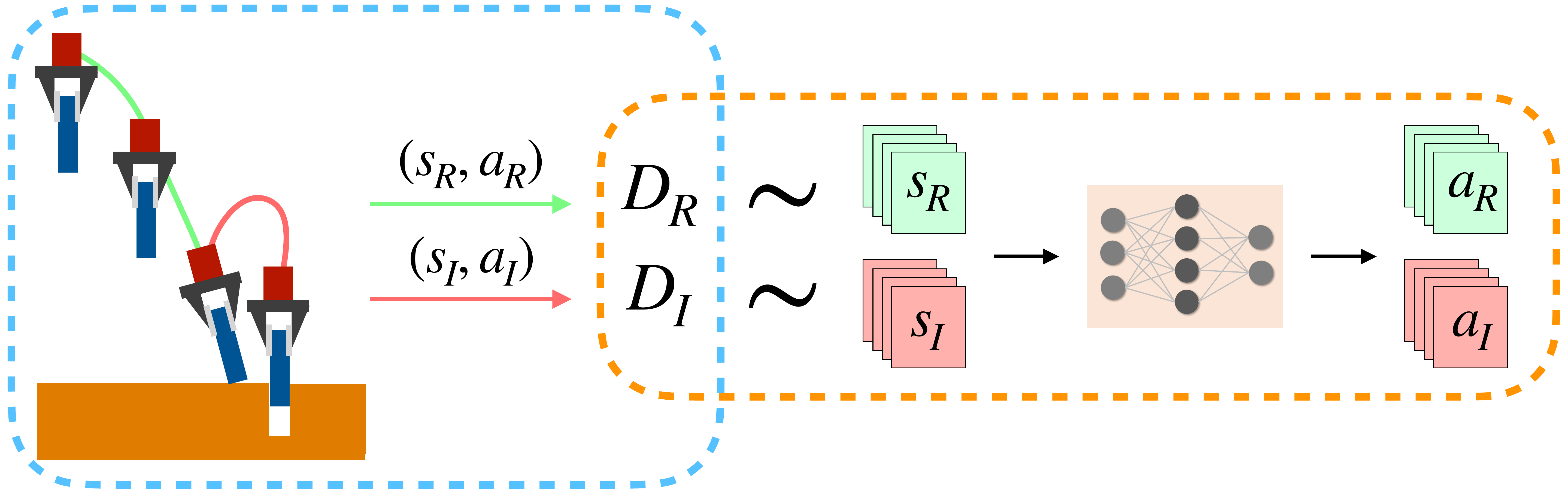}
    \caption{\textbf{\algoFull.} Every intervention trajectory (blue box) consists of portions where the policy was controlling the robot (green) and portions where the human was intervening (red). These are aggregated to separate datasets. The intervention portions usually start in the neighborhood of bottlenecks that require a structured sequence of actions to traverse successfully and they demonstrate how to traverse them. During training (orange box), we sample equal size batches from the non-intervention and intervention datasets and train with a behavioral cloning loss. Sampling equal size batches from each dataset re-weights the data distribution to reinforce intervention actions that demonstrate bottleneck traversal, while the non-intervention samples provide regularization that keeps the policy close to previous policy iterates.}
    \label{fig:method_overview}
    \vspace{-15pt}
\end{figure}

\section{Preliminaries}
\label{sec:problem}

We formalize the problem of solving a robot manipulation task as an infinite-horizon discrete-time Markov Decision Process (MDP), $\mathcal{M} = (\mathcal{S}, \mathcal{A}, \mathcal{T}, R, \gamma, \rho_0)$, where $\mathcal{S}$ is the state space, $\mathcal{A}$ is the action space, $\mathcal{T}(\cdot | s, a)$, is the state transition distribution, $R(s, a, s')$ is the reward function, $\gamma \in [0, 1)$ is the discount factor, and $\rho_0(\cdot)$ is the initial state distribution. At every step, an agent observes $s_t$, uses a policy $\pi$ to choose an action, $a_t = \pi(s_t)$, and observes the next state, $s_{t+1} \sim \mathcal{T}(\cdot | s_t, a_t)$, and reward, $r_t = R(s_t, a_t, s_{t+1})$. The goal is to learn an policy $\pi$ that maximizes the expected return: $\mathbb{E}[\sum_{t=0}^{\infty} \gamma^t R(s_t, a_t, s_{t+1})]$. 


We now review some methods for learning from demonstrations. Behavioral Cloning (BC)~\cite{pomerleau1989alvinn} is a common and simple method for learning from a set of demonstrations $\mathcal{D}$. It trains a policy $\pi_{\theta}(s)$ to clone the actions in the demonstrations with the objective: $\arg\min_{\theta} \mathbb{E}_{(s, a) \sim \mathcal{D}} ||\pi_{\theta}(s) - a||^2$.

Policies trained with BC often suffer from covariate shift, because small action errors can result in a state visitation distribution $\rho_{\pi}(s)$ that differs from the one provided in the demonstration data $\rho_{\mathcal{D}}(s)$. To address this issue, Ross et al.~\cite{ross2011reduction} introduced \textsc{DAgger}, which optimizes the objective $\arg\min_{\theta} \mathbb{E}_{s \sim \rho_{\pi_{\theta}}(s)} ||\pi_{\theta}(s) - \pi_D(s) ||^2$, where $\pi_D$ is the demonstrator policy. This objective ensures that the policy imitates the demonstrator policy on its own induced distribution of states, instead of the demonstrator's distribution of states. To optimize this objective, \textsc{DAgger} alternates between collecting state samples using the current policy iterate $(s, \pi_{\theta}(s))$, relabeling the states visited using the demonstrator policy $(s, \pi_D(s))$, and updating the policy with BC. 

However, relabeling all states in an offline manner is not feasible for humans in realistic continuous control tasks. Kelly et al.~\cite{kelly2019hg} introduced \textsc{HG-DAgger}, a simple variant of \textsc{DAgger} that does not require an explicit offline relabeling process. Instead, the demonstrator is allowed to intervene and control the agent during agent execution when they are unsatisfied with the agent performance. Thus, state samples are collected under a mixture policy $\pi(s) = G_H(s)\pi_H(s) + (1 - G_H(s))\pi_{\theta}(s)$, where $G_H(s)$ represents a gating function that corresponds to when the human decides to intervene. The intervention samples, $\{(s, a) | G_H(s) = 1\}$, are treated as the relabeled samples, and are added to the dataset in order to train the agent, while the on-policy agent samples are discarded. However, training the agent for the next round without on-policy agent samples can cause agent behavior to change significantly due to covariate shift, especially in contact-rich manipulation settings as explored in this work. To alleviate this issue, we develop a method that includes on-policy agent samples as regularization, but prioritizes intervention samples that demonstrate successful bottleneck traversal.


\section{Leveraging Remote Human Interventions}
\label{sec:method}

\begin{figure*}[!t]
    \centering
    \begin{subfigure}{0.8\textwidth}
       \centering
    \includegraphics[width=\columnwidth, trim=0 3.1cm 0 0, clip]{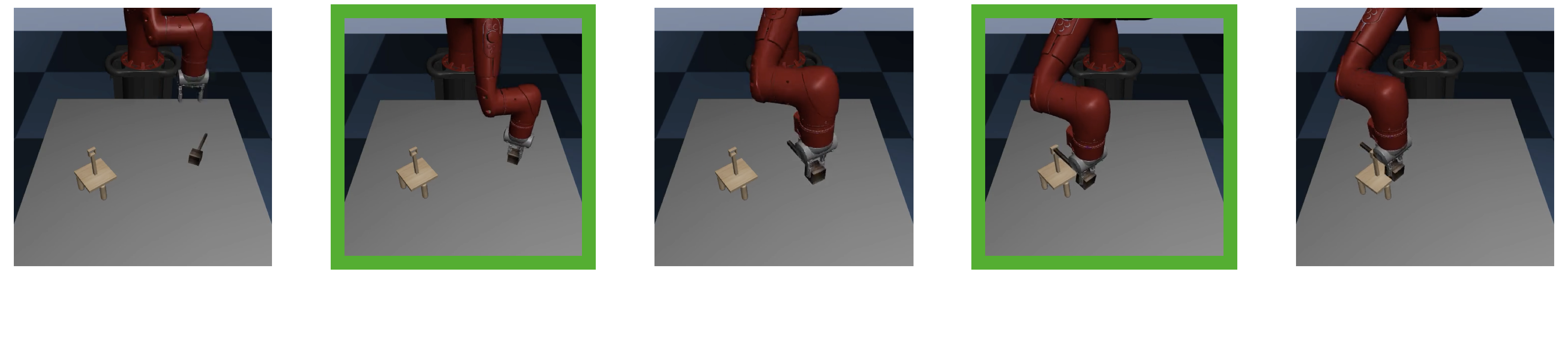}
    \caption{Threading} 
    \end{subfigure}
    \begin{subfigure}{0.8\textwidth}
       \centering
    \includegraphics[width=\columnwidth, trim=0 3.1cm 0 0, clip]{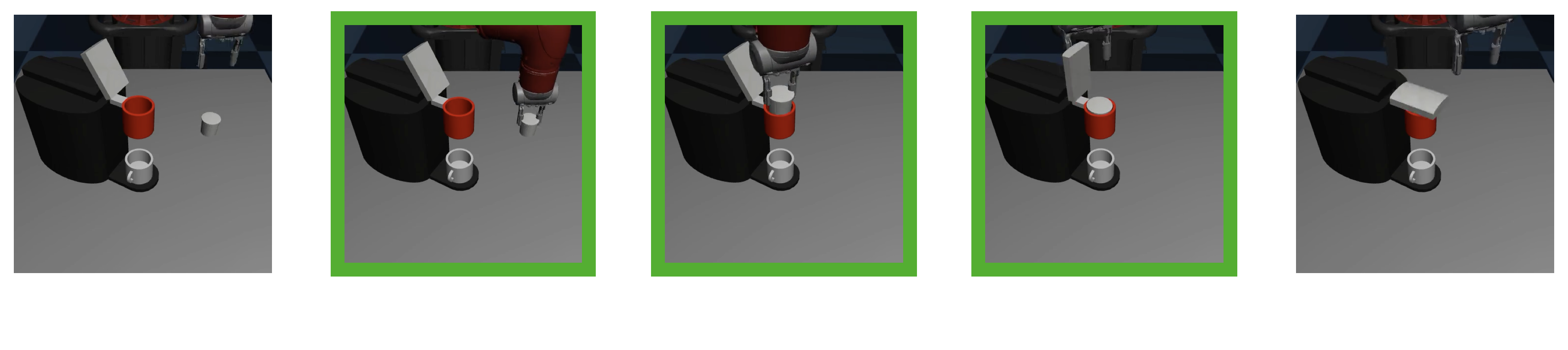}
    \caption{Coffee Machine} 
    \end{subfigure}
    \caption{\textbf{Tasks:} a) The Threading task requires the robot to carefully grasp a rod and insert it into the hole. It contains two bottleneck regions: grasping and insertion (green frames). b) The Coffee Machine task requires the robot to pick up a coffee pod, insert it into the machine, and then close the lid. It contains three bottleneck regions: grasping, insertion, and closing (green frames).
    }
    \label{fig:tasks}
    \vspace{-7pt}
\end{figure*}

We first describe our data collection system that allows remote operators to collect intervention data for manipulation tasks. Next, we discuss how learning from intervention data can be viewed as performing reinforcement learning where the human curates the dataset for the policy to learn from. Finally, we present \algoFull, an effective method for learning from intervention data, that leverages this insight to re-weight the dataset distribution according to the human intervention timings.

\subsection{Remote Teleoperation System}
We develop our data collection system on top of RoboTurk~\cite{mandlekar2018roboturk, mandlekar2019scaling} to allow remote operators to monitor policy execution and intervene when necessary. RoboTurk is a platform that allows users to collect task demonstrations through low-latency teleoperation. Users log on to a website that streams video from the robot workspace, and use a smartphone as a motion controller to control the robot. The platform works for both simulated and real robots, although in this work we focus on simulated domains\footnote{Our intervention system can be applied \textit{as-is} to real-world robots, but we were unable to access our robot arms due to current COVID measures.}. The robot simulation runs in remote servers hosted in the cloud to make it simple for users to participate in data collection. They only require a smartphone and a web browser to stream video.

We extend RoboTurk to enable remote users to monitor trained policies and intervene when necessary to help them improve (see Fig.~\ref{fig:system}). The user is able to pause and resume policy execution by tapping a button on their smartphone. The user can intervene by holding another button down and moving their phone in free space to apply relative pose commands to the robot arm. 

This provides a natural and simple way for users to intervene and apply corrections, since users can control the robot by moving or rotating their phone in a particular direction relative to the current robot pose. For example, the user can simply push their phone forward to make the arm move back, or twist their phone along a particular axis to apply the same relative rotation to the robot arm. Each recorded task demonstration now consists of a mix of human and policy samples. We record each completed task demonstration with binary intervention labels, and leverage the datasets collected by our system for imitation learning.

\subsection{Intervention-Based Policy Learning}

We first show how we can formulate intervention-based policy learning as an alternating optimization problem between the human and the policy. We then show that our algorithm emerges via assumptions on how the human solves this optimization problem. We start with the RL objective, which is to find a policy $\pi_{\theta}$ that maximizes the expected return $\mathbb{E}_{p_{\pi_{\theta}}(\tau)}[R(\tau)]$, where $\tau$ is a trajectory of states and actions, ${p_{\pi_{\theta}}(\tau)}$ is the distribution of trajectories induced by policy $\pi_{\theta}$, and $R(\tau)$ is the discounted return of the trajectory. We instead choose to maximize the log of the return, and then introduce a variational trajectory distribution $q(\tau)$ and a corresponding variational lower bound~\cite{abdolmaleki2018maximum}
\begin{align}
    \label{eq:elbo}
    J(\theta, q) = \mathbb{E}_{q(\tau)}[\log R(\tau) + \log p_{\pi_{\theta}}(\tau) - \log q(\tau)]
\end{align}

In our setting of intervention-based learning, we view $q(\tau)$ as a dataset distribution that is curated by the human. Eq.~\eqref{eq:elbo} can be maximized via Expectation-Maximization~\cite{dayan1997using}, which alternates between optimizing the trajectory distribution $q$ while holding the policy fixed and optimizing the policy parameters $\theta$ while holding $q$ fixed. Each round of intervention data collection and policy training corresponds to an iteration of EM. During each round, the human tries to maximize Eq.~\eqref{eq:elbo} by improving the dataset sample distribution $q(\tau)$ via interventions. This means that the human optimizes $q(\tau) = \arg\max_q J(\theta, q)$, which can be written as 
\begin{align}
    \label{eq:q_opt}
    q(\tau) = \arg\max_q \, \mathbb{E}_{q(\tau)}[\log R(\tau)] - KL[q(\tau) \,||\, p_{\pi_{\theta}}(\tau)]
\end{align}
where KL denotes the Kullback-Leibler divergence between the variational trajectory distribution and the one induced by the current policy. A human optimizes this objective by choosing to intervene in different regions of the state space to improve the task success rate of the trajectories in the dataset. Notice that the KL penalty in Eq.~\eqref{eq:q_opt} is implicitly encouraged in intervention data collection because all samples are on-policy, except for the human intervention samples.

Then, the base policy for the next iteration is trained by solving $\theta = \arg\max_{\theta} J(\theta, q)$ over the dataset curated by the human, which corresponds to the following objective
\begin{align}
    \label{eq:theta_opt}
    \theta = \arg\max_{\theta} \mathbb{E}_{(s, a) \sim q(\tau)} [\log \pi_{\theta}(a | s)]
\end{align}
If using a deterministic policy, Eq.~\eqref{eq:theta_opt} reduces to the BC loss $\arg\min_{\theta} \mathbb{E}_{(s, a) \sim q(\tau)} ||\pi_{\theta}(s) - a||^2$, where $q(\tau)$ is curated by the human during intervention data collection. Notice that the density $q(\tau)$ effectively assigns importance to each state-action sample, and is equivalent to weighting the BC loss on a per-sample basis. A number of works have leveraged this insight to develop RL algorithms~\cite{peters2007reinforcement, neumann2009fitted, abdolmaleki2018maximum, peng2019advantage}.

Thus, we can view intervention-based policy learned as optimizing a lower bound on the reinforcement learning objective via EM, where each iteration consists of a human curating a data distribution, and a policy update given by BC over the curated data distribution. In the next section, we discuss how this perspective can be leveraged to make better use of human interventions in manipulation tasks.

\subsection{Intervention Weighted Regression}

Different assumptions for how the human carries out the optimization in Eq.~\eqref{eq:q_opt} to produce the data distribution $q(\tau)$ result in different algorithms. A simple assumption is that the human specifies $q(\tau)$ directly via the intervention signal -- every sample where the human is intervening and controlling the robot is included in $q$ and the rest of the on-policy samples are discarded and not used for learning -- this is what \textsc{HG-DAgger}~\cite{kelly2019hg} does. However, excluding on-policy samples in the dataset distribution can cause the policy trained in the next round to change substantially from the current policy, and induce a significantly different distribution of policy failures due to covariate shift. The optimization in Eq.~\eqref{eq:q_opt} captures this intuition through the KL penalty, which rewards the $q$ distribution for including samples close to the policy distribution of states and actions.

Thus, we instead assume that the human-curated distribution $q(\tau)$ includes both intervention samples $\mathcal{D}_I$ and on-policy samples $\mathcal{D}_R$. However, $q$ also needs to assign density to each sample, as mentioned earlier, samples with higher density correspond to higher importance. 


Fortunately, the human intervention timings can be indicative of the importance of samples in manipulation domains. This is because robot manipulation tasks often contain \textit{bottlenecks}, which are regions of the state space that require a structured sequence of actions to traverse successfully (such as the task in Fig.~\ref{fig:pull}). Our key insight is that policies trained on offline demonstrations are most likely to incur errors near bottlenecks, and consequently, the timing of the human interventions identify the location of these bottlenecks and how to traverse them. Actions that successfully traverse bottlenecks should be \textit{reinforced} over others in a trajectory. 

The dataset collected via human interventions consists of a set of intervention data samples, $\mathcal{D}_I$, and on-policy data samples collected by the robot, $\mathcal{D}_R$. We assume that the human is specifying $q(\tau)$ by re-weighting the distribution of data with a parameter $\alpha$ such that $$q(s, a) \propto \alpha \rho_I(s, a) + \rho_R(s, a),$$ where $\rho_I(s, a)$ and $\rho_R(s, a)$ are the state-action distributions for the intervention and non-intervention samples respectively. The weight $\alpha$ corresponds to the amount of prioritization given to the intervention samples. 
We call our method \algoFull (\algoName) to reflect how reinforcement occurs by re-weighting the dataset distribution according to the user interventions.

In practice, we find that instead of tuning the value of $\alpha$ per dataset, choosing $\alpha = |\mathcal{D}_R| / |\mathcal{D}_I|$ so that $q(s, a)$ samples from $\rho_I(s, a)$ and $\rho_R(s, a)$ in equal proportion performs well across datasets (see Fig.~\ref{fig:method_overview}). This is equivalent to behavioral cloning with the objective $$\theta = \arg\min_{\theta} || \pi_{\theta}(s_I) - a_I||^2 + ||\pi_{\theta}(s_R) - a_R||^2$$ with $(s_I, a_I) \sim \mathcal{D}_I$, $(s_R, a_R) \sim \mathcal{D}_R$. 
\section{Experiments}
\label{sec:exp}


\begin{table*}[t!]
\centering
\parbox[t][][t]{.49\linewidth}{
\centering
\caption{Single-Operator Results on the Threading Task}
\begin{tabular}{l|ccc}
\hline
\textbf{Model}      & \textbf{Round 1}                  & \textbf{Round 2}                  & \textbf{Final}           \\ \hline
Base       & -                        & -                        & $58.0 \pm 9.2$ \\
Full Demos & -                        & -                        & $76.7 \pm 2.3$ \\
\textsc{HG-DAgger}  & $57.3 \pm 9.5$          & $62.7 \pm 5.0$          & $75.3 \pm 8.1$ \\
\algoName-NB   & $76.0 \pm 6.9$          & $72.0 \pm 3.5$          & $74.7 \pm 1.2$ \\
\algoName (Ours)       & $\mathbf{84.0 \pm 5.3}$ & $\mathbf{90.7 \pm 3.1}$ & $\mathbf{87.3 \pm 5.0}$ \\ \hline
\end{tabular}
\label{table:single}
}
\hfill
\parbox[t][][t]{.49\linewidth}{
\centering
\caption{Multi-Operator Results on the Coffee Machine Task}
\begin{tabular}{l|ccc}
\hline
\textbf{Model}      & \textbf{Round 1}                  & \textbf{Round 2}                  & \textbf{Final}                    \\ \hline
Base       & -                        & -                        & $52.0 \pm 3.5$          \\
Full Demos & -                        & -                        & $64.9 \pm 8.3$          \\
\textsc{HG-DAgger}  & $70.2 \pm 15.3$          & $71.1 \pm 9.7$          & $69.6 \pm 10.1$          \\
\algoName (Ours)       & $\mathbf{79.6 \pm 8.9}$ & $\mathbf{79.5 \pm 11.7}$ & $\mathbf{87.5 \pm 9.4}$ \\ \hline
\end{tabular}
\label{table:multi}
}
\vspace{-5pt}
\end{table*}

In this section, we evaluate our system and method on challenging contact-rich manipulation tasks that require precise control. We seek to demonstrate that policies trained on data collected by our system can outperform policies trained on an equivalent amount of full human demonstration trajectories and that \algoName outperforms alternatives for learning from the intervention data.

\subsection{Tasks}
All tasks were designed using MuJoCo~\cite{todorov2012mujoco} and the robosuite framework~\cite{robosuite2020} (see Fig.~\ref{fig:tasks}). The workspace consists of a Sawyer robotic arm in front of a table. The arm is controlled using an Operational Space Controller~\cite{khatib1987unified}.

\textbf{Threading}: The robot arm must pick up a wooden rod and insert it into a wooden ring (Fig.~\ref{fig:tasks}). The location of the wooden rod and ring are randomized at the start of each episode. This task contains two bottlenecks - the grasping of the rod and the insertion into the ring. The insertion must be performed carefully - the wooden ring can move easily if the rod hits the ring.
The observation space consists of the end effector pose, gripper finger positions, and the poses of the wooden rod and ring.

\textbf{Coffee Machine}: The robot arm must pick up a pod, insert it into the holder, and then close the lid of the coffee machine (Fig.~\ref{fig:tasks}). The location of the pod is randomized at the start of each episode. This task has three bottlenecks - grasping, insertion, and closing. Both the pod grasping and insertion must be precise, as small errors will cause the pod to slip out of the hand, or fail to be inserted into the holder. 
The observation space consists of the end effector pose, gripper finger positions, the poses of the pod and pod holder, the hinge angle of the lid, and binary contact indicators between the pod, pod holder, and gripper fingers.

\subsection{Baselines}

\textbf{\algoName (Ours)}: At each round, intervention samples and on-policy samples are aggregated to different datasets. The datasets are sampled in equal proportion and policies are trained with BC on the balanced dataset samples.

\textbf{\algoName-NB}: Same as \algoName, but all dataset samples are added to one dataset. No dataset balancing takes place, and uniform sampling is used.

\textbf{HG-Dagger}: As in~\cite{kelly2019hg, ablett2020fighting} at each round of intervention data collection, only the samples where the user was intervening are added to the dataset, while the policy samples are discarded, and policies are trained with BC.

\textbf{Full Demos}: A human operator collects full task demonstrations instead of interventions. The trajectories are added to the dataset and policies are trained with BC.

\subsection{Experiment Details}

\begin{table*}[t!]
\begin{tabular}{cc}
\parbox[t][][t]{.49\linewidth}{
\centering
\caption{Single-Operator Comparison across Final Threading Datasets Collected by Each Method}
\begin{tabular}{l|ccc|}
\cline{2-4}
                                & \multicolumn{3}{c|}{\textbf{Final Dataset}}                                    \\ \hline
\multicolumn{1}{|l|}{\textbf{Model}}     & \textsc{HG-DAgger}                & \algoName-NB                & \algoName (Ours)            \\ \hline
\multicolumn{1}{|l|}{\textsc{HG-DAgger}} & $75.3 \pm 8.1$          & $72.0 \pm 5.3$          & $81.3 \pm 4.2$ \\
\multicolumn{1}{|l|}{\algoName-NB} & $80.0 \pm 1.4$          & $74.7 \pm 1.2$          & $86.0 \pm 4.0$  \\
\multicolumn{1}{|l|}{\algoName (Ours)}      & $\mathbf{87.3 \pm 6.4}$ & $\mathbf{84.7 \pm 6.4}$ & $\mathbf{87.3 \pm 5.0}$ \\ \hline
\end{tabular}
\label{table:single_cross}
}
&
\parbox[t][][t]{.49\linewidth}{
\centering
\caption{Multi-Operator Comparison across Final Coffee Machine Datasets Collected by Each Method}
\begin{tabular}{l|cc|}
\cline{2-3}
                                & \multicolumn{2}{c|}{\textbf{Final Dataset}}                  \\ \hline
\multicolumn{1}{|l|}{\textbf{Model}}     & \textsc{HG-DAgger}                & \algoName (Ours)                      \\ \hline
\multicolumn{1}{|l|}{\textsc{HG-DAgger}} & $69.6 \pm 10.1$          & $71.6 \pm 16.1$          \\
\multicolumn{1}{|l|}{\algoName (Ours)}      & $\mathbf{85.6 \pm 6.5}$ & $\mathbf{87.5 \pm 9.4}$ \\ \hline
\end{tabular}

\label{table:multi_cross}
}
\end{tabular}
\vspace{-5pt}
\end{table*}

We conducted two studies to demonstrate the utility of our system and our method for learning from intervention data. In the first, a single operator collected datasets on the Threading task. In the second, three human operators collected datasets on the Coffee Machine task. 

Each operator started with an initial dataset that consisted of 30 task demonstrations, and a base policy that was trained on that dataset. For each intervention-based method, the operator performed 3 additional rounds of data collection. During each round, the operator collected trajectories until the number of intervention data samples reached roughly $33\%$ of the initial dataset samples to ensure that all methods would be able to receive the same number of human-annotated samples at each round, regardless of base policy quality, and to be consistent with prior work~\cite{kelly2019hg}. After each round, the base policy for the next round was obtained by training for a fixed number of epochs.
For the Full Demos baseline, each operator collected a single dataset of human demonstration trajectories which had the same number of samples as the initial dataset.

All policies are 2-layer LSTMs with hidden size 100, trained on a sequence length of 10 using Adam optimizers~\cite{kingma2014adam}. To evaluate each method, policies were saved at a fixed rate and evaluated with 50 rollouts for each checkpoint. 
All success rates presented in each table reflect the maximum average success rate obtained by each run over all model checkpoints, for 3 training runs with different seeds. 

\subsection{Experiments}

\textbf{Does data collected using our intervention-based system improve task performance more than an equivalent amount of full demonstration samples?} We present results on the Threading datasets collected by a single operator in Table~\ref{table:single}. The results show that our method outperforms the Full Demos baseline by a significant margin ($87.3\%$ vs. $76.7\%$). This trend holds true even for the intermediate rounds, showing that intervention-based data collection can produce higher quality policies with fewer human-annotated samples. We also see that other intervention-based baselines do not necessarily improve upon the Full Demos performance - both \textsc{HG-DAgger} and \algoName-NB reach roughly the same level of performance as the Full Demos baseline. Only our method is able to consistently leverage intervention data to outperform the Full Demos baseline.

\textbf{Does using our method outperform baselines that learn from intervention data?} As shown in Table~\ref{table:single}, the results demonstrate that our method outperforms both the \textsc{HG-DAgger} baseline and the variant of our method without balancing consistently in each round on the Threading task.

\textbf{Are results consistent across multiple human operators?} We present results averaged across 3 different operators on the Coffee Machine task in Table~\ref{table:multi}. The results show that our method consistently outperforms the \textsc{HG-DAgger} baseline in each round, leading to an average final task performance of $87.5\%$ (over $35\%$ improvement over the original base policy), while the baseline reaches $69.6\%$ success (about $18\%$ improvement). Furthermore, the average Full Demos performance is $64.9\%$ (about $13\%$ improvement). Together, these results demonstrate the value of intervention-based data collection over collecting full human demonstrations, and intelligently leveraging both the human intervention and non-intervention samples for learning.

\textbf{Does our method outperform other methods on datasets that were collected using their intermediate policies?} We take the final aggregated datasets for each intervention-based method, and train policies on this dataset for all models. The results in Table~\ref{table:single_cross} and Table~\ref{table:multi_cross} demonstrate that our method consistently outperforms other baselines on their collected datasets and can reach a level of performance close to its own collected dataset. This suggests that other baselines do not fail purely due to lower quality data or worse base policies at each iteration, but due to the way they leverage the data.

\textbf{Is there value in doing more rounds of intervention data collection compared to collecting the same amount of data in a single round?} We had a single operator collect a large Round 1 intervention dataset with an equivalent number of intervention samples as was collected during all 3 rounds of intervention data collection, for both tasks. We found that our method did not exhibit a significant difference in final performance on the Threading task, and about $7\%$ lower in average success rate on the Coffee Machine task on the large intervention dataset, compared to performing iterative data collection. This suggests that longer tasks that contain more bottlenecks might benefit from iterative data collection more than shorter tasks, which makes sense - the number of potential mistakes that the policy can make increases with the number of task bottlenecks, and a single base policy may not sufficiently cover the space well.


\section{Conclusion}
\label{sec:conclusion}

We built a data collection system that allows remote operators to monitor trained policies and intervene when necessary to help the policy complete the task. We developed a simple and effective method to leverage such intervention data that reshapes the data distribution to prioritize bottleneck traversal via the timing of the human interventions, which is important in manipulation settings. We demonstrated that training an agent on intervention data with our method substantially outperforms other intervention-based baselines, and is more effective than training the agent with an equivalent number of full human demonstration trajectories. We showed that our results hold over multiple human operators and that our method can more effectively learn from intervention data even if other methods' base policies were used to collect and aggregate data. This makes our method ideal for crowdsourced settings~\cite{mandlekar2018roboturk, mandlekar2019scaling}, since we anticipate that data will be obtained from a variety of trained base policies and human operators. We plan to explore this in future work, as well as conduct data collection with physical robot arms.

\clearpage

{\footnotesize 
\section*{Acknowledgment}
We would like to thank Albert Tung and Josiah Wong for helping with data collection. Ajay Mandlekar acknowledges the support of the Department of Defense (DoD) through the NDSEG program. We acknowledge the support of Toyota Research Institute (``TRI''); this article solely reflects the
opinions and conclusions of its authors and not TRI or any other Toyota entity.
}

\renewcommand*{\bibfont}{\footnotesize}
\begin{flushright}
\printbibliography 
\end{flushright}

\end{document}